\documentclass[manuscript,screen]{acmart}
\AtBeginDocument{%
  \providecommand\BibTeX{{%
    \normalfont B\kern-0.5em{\scshape i\kern-0.25em b}\kern-0.8em\TeX}}}

\setcopyright{none}
\renewcommand\footnotetextcopyrightpermission[1]{} 

\theoremstyle{definition}

\usepackage{algorithm}
\usepackage{url}
\usepackage{longtable}
\usepackage{enumitem}
\usepackage{placeins}
\usepackage{algpseudocode}
\algrenewcommand\algorithmicrequire{\textbf{Input:}}
\algrenewcommand\algorithmicensure{\textbf{Output:}}
\usepackage{caption}
\usepackage{subcaption}
\usepackage{listings}
\usepackage{hyperref}
\usepackage{appendix}

\setcopyright{acmcopyright}
\copyrightyear{2022}
\acmYear{2022}
\acmDOI{XXXXXXX.XXXXXXX}

\acmConference[ACM FAccT '22]{Make sure to enter the correct
  conference title from your rights confirmation email}{June 21--24,
  2022}{Seoul, Korea}

\begin{document}

\title{Counterfactual Multi-Token Fairness in Text Classification}

\author{Pranay Lohia}
\email{plohia07@in.ibm.com}
\affiliation{%
  \institution{IBM Research AI}
  \country{India}
}
\email{plohia07@in.ibm.com}

\renewcommand{\shortauthors}{Pranay, et al.}

\begin{abstract}
The counterfactual token generation has been limited to perturbing only a single token in texts that are generally short and single sentences. These tokens are often associated with one of many sensitive attributes. With limited counterfactuals generated, the goal to achieve invariant nature for machine learning classification models towards any sensitive attribute gets bounded, and the formulation of Counterfactual Fairness gets narrowed. In this paper, we overcome these limitations by solving root problems and opening bigger domains for understanding. We have curated a resource of sensitive tokens and their corresponding perturbation tokens, even extending the support beyond traditionally used sensitive attributes like \textit{Age}, \textit{Gender}, and \textit{Race} to \textit{Nationality}, \textit{Disability}, and \textit{Religion}. The concept of Counterfactual Generation has been extended to multi-token support valid over all forms of texts and documents.
We define the method of generating counterfactuals by perturbing multiple sensitive tokens as \textbf{Counterfactual Multi-token Generation}. The method has been conceptualized to showcase significant performance improvement over single-token methods and validated over multiple benchmark datasets. The emendation in counterfactual generation propagates in achieving improved \textbf{Counterfactual Multi-token Fairness}.
\end{abstract}

\maketitle
\section{Introduction}
In current times when there is a growing demand to uncover measures and overcome discrimination in the real world, humans are re-analyzing and re-designing their algorithms to mitigate the \textit{unfairness} present in their machine-learning classification models. An important aspect that arises for these models is \textit{Counterfactual Fairness}, which requires similar model predictions for identical texts even though referencing different sensitive groups. These references are mainly in terms of tokens present in texts, which we refer to as \textit{sensitive tokens}. To solve this problem, we generate \textit{counterfactuals} by perturbing these sensitive tokens present in original texts and check whether prediction changes or not. By performing these tests, we can identify original texts and their counterfactuals where unfairness is present. These counterfactuals are then tagged the same labels as original texts and augmented with originals for re-training the model. This is done to make the classification model invariant of the sensitive tokens and achieve fairness. We have used the notion of Counterfactual given by \cite{garg2019counterfactual} where counterfactual doesn't necessarily result in changing the label by the classifier.

Past works have failed to make much progress in this area because of the limitations of resources involving sensitive tokens and their corresponding perturbations. Only a \textit{tip of the iceberg} set of sensitive attributes are studied. These sensitive attributes like age, gender, and race, etc. although addressed a lot in literature, don't have a resource of tokens and their perturbations to make their study more standardized. Sensitive attributes like \textit{Disability} and \textit{Religion} which demands much importance, have just been skimmed through in some literature. Algorithms and methods developed to address \textit{Counterfactual Fairness} have validated their applications on shorter texts or texts involving only one sensitive token. Examples are: 1. \textit{Some people are gay}, 2. \textit{I saw a black person} \cite{garg2019counterfactual} \cite{disabilitywords}. We can find only a single sensitive token i.e., \textit{gay} and \textit{black} present in each text respectively. However, much of real-world texts are not such small or single sentences. They are big sentences and paragraphs involving multi-token i.e., multiple sensitive tokens of the same sensitive attribute or tokens spread across multiple sensitive attributes.

Example: \textit{She is a Christian, who visits Church.} Sensitive tokens are \textit{She, Christian, and Church}.

Apart from perturbing them individually using single-token methods, we can also generate more counterfactuals by perturbing all three tokens together or in pairs e.g., \textit{He is an obedient Jew, who visits Synagogue daily} and \textit{She is an obedient Hindu, who visits Temple daily}. We can generate more counterfactuals by perturbing multiple tokens. A higher flip-rate \cite{black2020fliptest} will signify that we can identify more unfairness present in the model. We formulate this method of generating counterfactuals by perturbing multiple sensitive tokens in texts as Counterfactual Multi-token Generation. Using the augmented data generated, we can make the model achieve better fairness, which we define as Counterfactual Multi-token Fairness.

We make the following four major contributions: 
\begin{enumerate}
    \item Data curation of words and identification of their corresponding perturbations for sensitive attributes age, disability, race, nationality, gender, and religion.
    \item Identification of sensitive attributes multi-tokens in texts and proposing a better performing method of generating counterfactuals i.e., Counterfactual Multi-token Generation than the single-token perturbation method.
    \item Employing the usage of \textit{Explainability} in improving the performance of the counterfactual generation and fairness method.
    \item With the application of the Counterfactual Multi-token generation method in data augmentation based bias-mitigation, we define a significantly high-performing Counterfactual Fairness method i.e., Counterfactual Multi-token Fairness.
\end{enumerate}

The remainder of the paper is organized as follows. First, we provide a study of related works in Sec.2. In Sec.3, we provide the definition of Counterfactual Multi-token Generation followed by the algorithmic details in Sec.4. In Sec.5, we propose Counterfactual Multi-token Fairness. In Sec.6, we provide a detailed resource of sensitive words/tokens and their corresponding perturbation tokens. In Sec.8.1, we showcase performance improvements of the multi-token method over the standard single-token method. The usage of explainability in improving fairness is addressed in Sec.8.2. In Sec.8.3, we validate how our generation method helps in achieving overall improved Counterfactual Multi-token Fairness. Finally, we conclude the paper in Sec.9 with future work propositions.

\section{Related Work}

\textbf{Fairness in Text.} Fairness and Ethics literature has a deep focus on classification model fairness whether detection or mitigation. \cite{DBLP:journals/corr/abs-1810-01943} articulates various metrics and algorithms to handle fairness in machine learning models. Training data can contain biases because of historical prejudices, and these can propagate to machine learning models. With the advancement in AI and much of the technologies driven towards using AI models, the presence of unwanted prejudices can hamper the reputation both socially and economically. In terms of modalities, structured data and text is being studied thoroughly in fairness literature. \cite{bolukbasi2016man}, \cite{dwork2012fairness}, \cite{hardt2016equality}, \cite{louizos2015variational}, \cite{zemel2013learning} lays down foundational areas for studying Fairness in Text. \cite{bolukbasi2016quantifying}, \cite{netzer2019words}, \cite{zhao2018learning}, \cite{gonen2019lipstick}, \cite{zhao2019gender}, and \cite{yang2020causal} provides methods for removing unfairness from word representations and embeddings. \cite{shandilya2018fairness} talks about fairness in text summarization. 

\noindent
\textbf{Counterfactual Multi-token Generation.} Sensitive attributes based debiasing has been studied, but are limited to small sentences involving single-token perturbation handling. \cite{prost2019debiasing}, \cite{bordia2019identifying}, \cite{sun2019mitigating}, \cite{font2019equalizing}, and \cite{qian2019reducing} highlights mitigating gender bias in machine translations and natural language processing. \cite{manzini2019black} talks about unfairness in AI models when announcing judgments on grounds of persons' race. Table \ref{table:paper} lists the approximate number of papers or studies associated with various sensitive attributes. \cite{prabhakaran2019perturbation} talks about how sensitive tokens can be perturbed to identify and mitigate unfairness in machine learning models. Removal of unintended bias from sentiment analysis and toxicity detection models has been studied in \cite{caliskan2017semantics}, \cite{webster2018mind}, \cite{agelist}, and \cite{de2019bias}. \cite{kiritchenko2018examining}, \cite{may2019measuring}, and \cite{gonen2019lipstick} talks about how counterfactuals or perturbation generation can help in removal of unfairness from unstructured texts. \cite{kusner2017counterfactual}, \cite{ribeiro2020beyond}, and \cite{kilbertus2017avoiding} forms the basis of our work on Counterfactual Multi-token Generation and Fairness. They provide a formal definition of counterfactual fairness and counterfactual token fairness. \cite{garg2019counterfactual} highlights how model prediction changes when identity terms involving sensitive tokens, in it predominately gender and sexual orientation tokens (such as gay, lesbian, transgender
etc.), are perturbed. 

\begin{table}[h!]
\small
\makebox[\linewidth]{
\begin{tabular}{|c|c|}
    \hline
    \textbf{attribute} & \textbf{\#papers associated}\\
    \hline
    age&33\\
    disability&5\\
    race&35\\
    gender&42\\
    religion&3\\
    \hline
\end{tabular}
}
\caption{\label{table:paper} Approximate number of papers or studies associated with sensitive attributes}
\end{table} 

\noindent
\textbf{Counterfactual Multi-token Fairness.} Social scientists have raised concerns over the entire concept of Counterfactual Fairness \cite{kohler2018eddie} \cite{krieger2014causal}. However, with the growing misconduct against minorities in the real-world, the need for changing the ever-existing social-construct and prejudice need to be broken. \cite{kusner2017counterfactual} points out that fairness is achieved only when original and counterfactuals have the same prediction under some constraints \cite{dwork2012fairness}. \cite{wachter2017counterfactual} and \cite{grari2020adversarial} characterize adversarial examples similar to counterfactuals, where similar output is expected as in original texts. \cite{DBLP:conf/ijcai/Wu0W19}, \cite{kilbertus2020sensitivity}, and \cite{di2020counterfactual} validates how counterfactuals fairness is important in achieving model fairness. Explainability methods like LIME \cite{ribeiro2016should} and Anchors \cite{ribeiro2018anchors} suggest that explainability token(s) anchors the predictions locally. These tokens are responsible for the model's prediction. Since perturbations to these tokens can alter model predictions, they can be combined with sensitive tokens to form the multi-token setup which can be perturbed to their corresponding perturbations to generate counterfactuals. Machine learning literature proposes multiple methods to achieve fairness, however, data augmentation involving perturbation of tokens to generate counterfactuals is the most viable in achieving Counterfactual Fairness \cite{park2018reducing} when handling multiple sensitive attributes. These flipped counterfactuals \cite{garg2019counterfactual} are augmented with the original data and are used to re-train the model. This makes the model invariant to the sensitive tokens and their corresponding perturbations, making the model counterfactually fair \cite{garg2019counterfactual}.

\section{Problem Definition}
In \cite{kusner2017counterfactual} and \cite{garg2019counterfactual}, counterfactuals are generated but limited to perturbing single token in the text. We term this type of counterfactuals generation as Counterfactual Single-token Generation.
E.g., \textbf{Original Text:} \{\textit{Who would read a book by a woman.}\} will be perturbed to, \textbf{Perturbed Text:} \{\textit{Who would read a book by a man.}\}\\
There are four limitations to the above technique:
\begin{enumerate}
    \item Single token counterfactuals miss the contextual information leading to unrealistic sentences \cite{boleda2016lambada}. e.g., Original Text: \textit{The statue of Christ is magnificent in the church.}
    will be perturbed to Either: \textit{The statue of Vishnu is magnificent in the church.}, Or: \textit{The statue of Christ is magnificent in the temple.} Just from the observation, we can say that these are contextually unrealistic and semantically wrong perturbations.
    \item It doesn't generate correct counterfactuals when the text has multiple token(s) of a particular sensitive attribute. e.g., \textit{He and his friends are too jovial.} Here, one needs to choose between \textit{He} and \textit{his} for performing counterfactual single-token generation, which not only misses other sensitive words/tokens for perturbation but makes the perturb text grammatically and syntactically wrong.
    \item It doesn't have support for counterfactuals generation involving a combination of multiple sensitive attributes. e.g., \textit{History of our nation only started when the white man arrived.} Based on single token generation, we have to either choose between \textit{white}, which belong to \textit{race} attribute and to be perturbed to \textit{black}, or \textit{man}, which belong to \textit{gender} attribute and to be perturbed to \textit{woman}. While these are realistic counterfactuals, but one misses the third counterfactual where multiple tokens are perturbed together. Multi-token Counterfactual: \textit{History of our nation only started when the black woman arrived.}
    \item Since we are missing important counterfactuals, our analysis of Counterfactual fairness \cite{kusner2017counterfactual} gets flawed. There can be many texts whose multi-token perturbation counterfactuals will get a different label than their original, and will not be covered in the analysis. Past counterfactuals generation methods don't take into consideration whether the counterfactuals generated are factually correct or not.
\end{enumerate}

\subsection{Counterfactual Multi-token Generation} 
To overcome these limitations and attain superior performance, we propose a multi-token(s) based counterfactual generation method as following:

\textit{Notation 3.1.} For a text input $x$ $\epsilon$ $X$, where $X$ is the text data and $x$ is a sequence of tokens [$x_1$, $x_2$, $x_3$,.., $x_n$]. Using a classifier $f$, we can predict $f(x)$. \cite{garg2019counterfactual} defines set of counterfactuals associated with $x$ as $\Phi(x)$. We denote set of sensitive attributes as $S$ = $A, B, C, D,..,etc$. Representation of any sensitive attribute, for e.g., $A$, in the text input is done by its associated words which we term as \textit{sensitive multi-token(s)}, i.e., $\{a_1, a_2,..,a_n\}$. Counterfactuals are generated by perturbing these sensitive multi-token(s), for e.g., $\{a_1, a_2,..,a_n\}$, to their corresponding perturbations, i.e., $\{a_1', a_2',..,a_n'\}$. Hence, sensitive multi-token(s) belonging to these sensitive attributes are denoted as $s$ = $[\{a_1, a_2,..,a_n\}, \{b_1, b_2,..,b_n\},...etc.]$, and their corresponding perturbations as: \\ $s'$ = $[\{a_1', a_2',....,a_n'\}, \{b_1', b_2',....,b_n'\},...etc.]$. 

\begin{definition} \label{gen_def}
Generation of counterfactuals by perturbing every sensitive token(s) present in the text both individually and in various combinations. Using the above notations, we formulate \textbf{Counterfactual Multi-token Generation} as:
\begin{equation}
    \label{eqn:gen_formula}
    \Phi_S(x) = \bigcup_{s\neq s' \epsilon S}^{} \Phi_{s,s'}(x)
\end{equation}

The above formulation is preceded by identification of all sensitive token(s) present in the text. We denote identified sensitive token(s) as $x_s$, where 
\begin{equation}
    \label{eqn:identification_formula}
    x_{s} = \bigcup_{x_{i} \epsilon s}^{} x_{i}, \forall x_{i} \epsilon x
\end{equation}

After identification, sensitive token(s) $x_s$ are replaced with their corresponding perturbations $x_s'$ in the text $x$ using the method proposed in \cite{bolukbasi2016man}. We get a direct mapping between sensitive multi-token(s) $s$ and their corresponding perturbations $s'$. This helps in generating semantically correct counterfactuals. The perturbations are performed both individually and in various combinations of the sensitive token(s) to generate multiple counterfactuals $\Phi_S(x)$ using the multi-token generation formulation as in the equation \ref{eqn:gen_formula}.
\end{definition}

\section{Algorithm} \label{gen_method}

We present the algorithmic details associated with generating counterfactuals using the multi-token generation method. It involves the identification of sensitive token(s), followed by their replacement with perturbation token(s) in various combinations keeping in mind semantic, syntactic, and dependency information.

\subsection{Identification of sensitive attributes' words or token(s) in text}
The presence of words or token(s) in each text related to age, disability, race, nationality, gender, and religion is noted. Table \ref{table:texttopicwords} showcases certain examples, where each text can have multiple sensitive attributes' multi-token either belonging to a particular sensitive attribute or of multiple sensitive attributes combined i.e., age, disability, race, nationality, gender, and religion.
\begin{table*}[h!]
\centering
\small
\makebox[\linewidth]{
\begin{tabular}{|p{5cm}|p{2cm}|p{1cm}|p{1.5cm}|p{1.5cm}|p{1cm}|p{1.5cm}|}
    \hline
    \textbf{Text} & \multicolumn{6}{|c|}{\textbf{Multi-token}} \\
    \cline{2-7}
    &\textbf{age} & \textbf{disability} & \textbf{race} & \textbf{nat} & \textbf{gender} & \textbf{religion} \\
    \hline
    Love it! Alas, though, there are no fun non-derogatory nouns for the male equivalent of tomboy. My son's choices will be relegated to "sissy"/"mama's boy" or "variant"/"nonconforming". Yes, he's a "boy" and "human" -- but as you noted, the power of self-identification as something cool and different can be magical.& & & & & boy, male, son, he& \\
    \hline
    Interesting take on religion Jeff, thanks.  I remember all those "devout, born-agains" just loving young Dubya because he was--like them--a "good Christian man."  Who laughed as he sent a mentally-ill white woman to the gas chamber as governor of Texas. A lying master of dirty tricks, who fled the Catholic church, in Texas, for one with more voters.  Most Europeans have gotten past tying religion and politics.  If only we could.  God may not be dead, but she certainly has better things to do than hang out in the voting booth.&young&mentally, ill&white&&his, man, she, he, woman&christian, church, god\\
    \hline
    He has said in public about Mexicans, Muslims, women, blacks, handicapped, and so on. I don't hate him, but I do seriously object to his candidacy. Furthermore, hating an individual can't be bigotry. Bigotry is about hating groups and collections of people for their characteristics, gender, background, color, religion, preferences, age, etc. Disliking a person, even hating that person, for what he or she says, believes, or does, rather than that person's classification, is not bigotry.&handicapped&&black&mexicans&his, women, she, him, he&muslims\\
    \hline
\end{tabular}
}
\caption{\label{table:texttopicwords}Sensitive attributes' multi-token(s) present in text}
\end{table*} 

\subsection{Counterfactual Multi-token Generation method}
 Counterfactuals are generated by replacing identified sensitive token(s) with their corresponding perturbation token(s) individually and in various combinations. From the definition \ref{gen_def}, it is clear that multi-token method of counterfactual generation is super-set of single-token generation method \cite{kusner2017counterfactual}. Popular grammar checker \cite{grammar} and language models \cite{peters2018deep, devlin2018bert, brown2020language} are used to filter out unrealistic counterfactuals. Table \ref{table:countgen} showcases some examples based on \textit{Counterfactual Multi-token Generation}.
\begin{table}[h!]
\small
\makebox[\linewidth]{
\begin{tabular}{|p{3cm}|p{3cm}|}
    \hline
    \textbf{Original} & \textbf{Counterfactuals}\\
    \hline
    She is going to church. & He is going to church.\\
    & She is going to temple.\\
    & He is going to temple.\\
    \hline
    He and his friends are amazing. & She and her friends are amazing.\\
    \hline
    He has said in public about women. & She has said in public about women.\\
    & He has said in public about men.\\
    & She has said in public about men.\\
    \hline
\end{tabular}
}
\caption{\label{table:countgen}Examples of Counterfactual Multi-token generation}
\end{table}  

\subsubsection{Dependency parsing}
Dependency parsing is needed to segregate non-dependent multi-token during counterfactuals generation. It is only valid for multi-token setup.
Method:
\begin{enumerate}
    \item Dependency tree in each text is identified using spaCy \cite{spaCy} based dependency parser.
    \item Token(s) present in separate branches of the dependency tree are put in separate lists.
    \item Based on these lists, multiple counterfactuals are created for a particular text separately.
\end{enumerate}
For e.g., \textit{She is going to church, a white guy will be there too.} Based on dependency parsing, \textit{She is going to church} and \textit{a white guy will be there too} belongs to two different dependency trees. Since, both the clauses are independent, counterfactuals involving perturbations happening across both clauses together are redundant for the classification task. 
\newline
Hence, the useful counterfactuals are:
\begin{enumerate}

    \item\textit{He is going to church, a white guy will be there too.}
    \item\textit{She is going to temple, a white guy will be there too.}
    \item\textit{He is going to temple, a white guy will be there too.}
    \item\textit{She is going to church, a black guy will be there too.}
    \item\textit{She is going to church, a white girl will be there too.}
    \item\textit{She is going to church, a black girl will be there too.}
\end{enumerate}

Dependency parsing, Syntactical, and Semantic check are part of the post-processing steps performed after counterfactuals generation. 

\section{Counterfactual Multi-token Fairness}
\textit{Counterfactual Multi-token Generation} is followed by \textit{Counterfactual Bias} detection, method for improving the performance of the generation method using \textit{Explainability}, and finally data augmentation based bias-mitigation to achieve \textit{Counterfactual Multi-token Fairness}.

\subsection{Counterfactual Bias detection}
\textit{Counterfactual Bias} detection \cite{kusner2017counterfactual} involves identifying the original texts whose counterfactuals, generated after perturbation, produces a different label (or the label flips) than its original label. Machine-learning models are trained on the original samples/texts. For a particular dataset, the trained model is used to predict labels for counterfactuals. Counterfactuals are labeled as flipped if their predicted label is different from the predicted label of original texts. For analyzing \textit{Counterfactual Bias}, we define:

\begin{definition} 
\textbf{\textit{flip-rate}} as the ratio of the number of original texts whose at least one counterfactual flipped and the total number of original texts.
\end{definition}

Higher \textit{flip-rate} signifies that the method is able to identify more original texts which has \textit{Counterfactual Bias}. 
Counterfactuals of original texts having \textit{Counterfactual Bias} are then tagged labels of original texts. After tagging, all counterfactuals are concatenated together to form \textit{Counterfactual Data}. Original and counterfactual data are together used in achieving \textit{Counterfactual Multi-token Fairness}. An important inference to draw from it will be how a larger size of \textit{Counterfactual Data}, i.e., higher flip-rate, helps in achieving better fairness.

\subsection{Performance improvement using Explainability}
Based on the method articulated in section \ref{gen_method}, there is a possibility that in many cases, labels of counterfactuals may not flip because words identified may not be sufficient in contributing towards a particular label. This will bring us to a situation where we are deprived of an ample number of flipped counterfactuals. Also, in our task to achieve fairness, we are losing out on identifying several original texts which have \textit{Counterfactual Bias}. Counterfactuals of texts can be used in achieving a higher percentage of fairness. To solve this deficiency, we used standard model explanation methods like LIME \cite{ribeiro2016should} and Anchors \cite{ribeiro2018anchors}. These methods help in identifying token(s) that are contributing to the class label, and perturbing them will change the class label. These token(s) will also be indirectly dependent or related to the sensitive token(s). Hence, to generate more efficient counterfactuals and to achieve a higher percentage of \textit{Counterfactual Multi-token Fairness}, we need to combine explainability words with the sensitive token(s). With sensitive words as the default tokens to be used for perturbation, we append explainability words to the list if there is an intersection between sensitive words and explainability words. In counterfactuals, antonyms of explainability words are used as their perturbing words.
For e.g., Original Text: \textit{He is a polite white man.}. Sensitive token(s) are : \textit{He, white, man}, Explainability token(s) are: \textit{polite, man}.
Hence, new counterfactuals generated are:
\textit{1. She is a rude black woman.}
\textit{2. He is a rude black man.}
\textit{3. He is a rude white man.}
\textit{4. She is a rude white woman.}
Counterfactuals generated using the multi-token method are tagged the same label as the original and are termed as \textit{Counterfactual Data}.
They are augmented with the original data to form the model re-training set. This is done to make the model invariant of the counterfactuals. A significant decrement in flip-rate on the test data using the re-trained model showcases the model becoming fairer and invariant to the counterfactuals. We define this method of achieving fairness based on generating multi-token counterfactuals as \textbf{\textit{Counterfactual Multi-token Fairness}}.

\section{Sensitive Attributes' Words and Perturbations} \label{topic-words}
Counterfactuals are generated by perturbing sensitive words in the original text with their corresponding perturbations. Hence, there is a need for a resource of these words and perturbations for sensitive attributes. We have curated words and corresponding perturbations for the following sensitive attributes. The perturbations are identified by querying the antonyms of sensitive words from Merriam-Webster's Thesaurus \cite{merriam1978merriam}.

\subsection{AGE}
\subsubsection{Words Curation and Perturbations Identification}
To identify age-related words, we curate a list of age-related terms, as mentioned in \cite{agelist}. It categorizes words into two divisions i.e., \textit{old} and \textit{young}. Extending the list, we consider seed words in each division and extend them using distance in word embedding space. In Table \ref{table:mainageperturbationwords} complete resource of age-related words and their corresponding perturbations have been articulated.
\begin{table}[h!]
\centering
\small
\makebox[\linewidth]{
\begin{tabular}{|c|c|c|c|}
  \hline
  \textbf{Words} & \textbf{Perturbations} & \textbf{Words} & \textbf{Perturbations}\\  
  \hline
  parents & children &
  
  elders & kids\\
  
  elderly & kiddish &
  
  adult & youth\\
  
  old & teenage &
  
  elder & teenager\\
  
  parent & child &
  
  old & young\\
  
  older & younger &
  
  oldest & youngest\\
  
  young & old &
  
  children & parents\\
  
  kids & elders &
  
  kiddish & elderly\\
  
  teens & elders &
  
  youngster & adult\\
  
  teenage & old &
  
  teenager & elder\\
  
  teenagers & elders &
  
  child & parent\\
  
  younger & older &
  
  youngest & oldest\\
  \hline
\end{tabular}
}
\caption{\label{table:mainageperturbationwords}Words and their corresponding perturbations for sensitive attribute \textit{age}}
\end{table}

\subsection{DISABILITY}
\subsubsection{Words Curation and Perturbations Identification}
Social Biases as barriers for a person with disabilities have been studied in \cite{disabilitywords}. It highlights terms or words used in texts when representing disability. Disability-related words are queried, and their antonyms are identified as corresponding perturbations. Table \ref{table:maindisabilityperturbationwords} provides a complete resource for disability-related words and their corresponding perturbations.
\begin{table}[h!]
\centering
\small
\makebox[\linewidth]{
\begin{tabular}{|c|c|c|c|}
    \hline
    \textbf{Words} & \textbf{Perturbations} & \textbf{Words} & \textbf{Perturbations}\\  
    \hline
    mental & sane &
    
    mentally & intellectually\\
    
    illness & healthy &
    
    blind & sighted\\
    
    deaf & hearing &
    
    sight-deficient & sighted\\
    
    depression & cheerfulness &
    
    insane &  sane\\
    
    dyslexia &  unimpaired &
    
    slow-learner &  fast-leaner\\
    
    losers &  winners &
    
    loser & winner\\
    
    slow & fast &
    
    depressed & cheerful\\
    
    disorder & in-order &
    
    ill & healthy\\
    
    problems & solutions &
    
    issues & solutions\\ 
    
    help & no-help &
    
    treatment & no-treatment\\
    
    care & neglect &
    
    medication & hindrance\\
    
    therapy & inattention &
    
    treated & mistreated\\
    
    counseling & discourage &
    
    meds &    diseases\\
    
    medications & hindrance &
    
    homeless & settled\\
    
    drugs & diseases &
    
    homelessness & settleness\\
    
    drug & disease &
    
    alcohol & non-intoxicant\\
    
    police & civilian &
    
    addicts & non-addicts\\
    \hline
\end{tabular}
}
\caption{\label{table:maindisabilityperturbationwords}Words and their corresponding perturbations for sensitive attribute \textit{disability}}
\end{table}

\subsection{RACE}
\subsubsection{Words Curation and Perturbations Identification}
As a sensitive attribute, Race has been studied thoroughly in standard datasets like UCI \cite{Dua:2019} Adult (an income dataset based on a 1994 US Census database; sensitive attributes: sex, race), and ProPublica COMPAS (a prison recidivism dataset; sensitive attributes: sex, race). We have compiled race-related words from these two datasets in the following list: [\textit{black, white, asian, caucasian, hispanic, asian-pac-islander, amer-indian-eskimo, arabic, and oriental}]. Perturbations for \textit{Race} can be any of the other words present in their respective lists. Table \ref{table:raceperturbationwords} shows a subset of race related words and their corresponding perturbations from which one word can be randomly chosen as its perturbation for generating counterfactuals.

\begin{table}[h!]
\centering
\small
\begin{tabular}{|c|c|}\hline
        \textbf{Topic words} & \textbf{Perturbation words}\\
        \hline
        black & white,asian,caucasian,\\
        &hispanic,asian-pac-islander,\\
        &amer-indian-eskimo,arabic,oriental\\
        \hline
        white & black,asian,caucasian,\\
        &hispanic,asian-pac-islander,\\
        &amer-indian-eskimo,arabic,oriental\\
        \hline
        asian & black,white,caucasian,\\
        &hispanic,asian-pac-islander,\\
        &amer-indian-eskimo,arabic,oriental\\
        \hline
        British & Canadians,Indians\\
        &Americans,Chinese,other nationalities\\
        \hline
        Chinese & Canadians,Indians\\
        &Americans,Russians,other nationalities\\
        \hline
        Americans & Canadians,Indians\\
        &Chinese,Russians,other nationalities\\
        \hline
\end{tabular}
\caption{\label{table:raceperturbationwords}Words and their corresponding perturbations for sensitive attribute \textit{race}}
\end{table}

\subsection{NATIONALITY}
\subsubsection{Words Curation and Perturbations Identification}
Nationality: We have curated 222 nationality-related words from the Wikipedia list of nationalities \cite{wiki:NAT}. Perturbations for \textit{Nationality} can be any of the other words present in their respective lists. Table \ref{table:natperturbationwords} shows a subset of nationality related words and their corresponding perturbations from which one word can be randomly chosen as its perturbation for generating counterfactuals.

\begin{table}[h!]
\centering
\small
\begin{tabular}{|c|c|}\hline
        \textbf{Words} & \textbf{Perturbations}\\
        \hline
        British & Canadians,Indians\\
        &Americans,Chinese,other nationalities\\
        \hline
        Chinese & Canadians,Indians\\
        &Americans,Russians,other nationalities\\
        \hline
        Americans & Canadians,Indians\\
        &Chinese,Russians,other nationalities\\
        \hline
\end{tabular}
\caption{\label{table:natperturbationwords}Words and their corresponding perturbations for sensitive attribute \textit{nat}}
\end{table}

\subsection{GENDER}
\subsubsection{Words Curation and Perturbations Identification}
To identify gender words, we take the list described in \cite{bolukbasi2016man}. We have considered gender-specific keywords extracted using word embeddings described in \cite{bolukbasi2016man} totaling 165-word replacements. These word replacements are perturbations words. The subset resource containing gender words and their corresponding perturbations are articulated in Table \ref{table:maingenderperturbationwords}. The complete resource is detailed in the Appendix.

\begin{table}[h!]
\centering
\small
\makebox[\linewidth]{
\begin{tabular}{|c|c|c|c|}
    \hline
    \textbf{Words} & \textbf{Perturbations} & \textbf{Words} & \textbf{Perturbations}\\
    \hline
    
    he & she &
    he's & she's \\
    his & her &
    him & her \\
    man & woman &
    men & women \\
    spokesman & spokeswoman &
    wife & husband \\
    himself & herself &
    mother & father \\
    chairman & chairwoman &
    daughter & son \\
    guy & girl &
    boys & girls \\
    brother & sister &
    female & male \\
    dad & mom &
    actress & actor \\
    girlfriend & boyfriend &
    lady & gentleman \\
    businessman & businesswoman &
    grandfather & grandmother \\
    nephew & niece &
    gays & lesbians \\
    maiden & master &
    bisexual & heterosexual \\
    bachelor & bachelorette &
    prince & princess \\
    monks & nuns &
    lad & lass \\
    fiancee & fiance &
    maternal & paternal \\
    widows & widowers &
    motherhood & fatherhood \\
    witch & wizard &
    monk & nun \\
    queens & kings &
    housewives & househusbands \\
    patriarch & matriarch &
    pa & ma \\
    \hline
    
\end{tabular}
}
\caption{\label{table:maingenderperturbationwords}Words and their corresponding perturbations for sensitive attribute \textit{gender}}
\end{table}

\subsection{RELIGION}
\subsubsection{Words Curation and Perturbations Identification}
Till now religion as a sensitive attribute hasn't been analyzed in depth. One of the novel aspects of this paper is the identification of religion-related words and their corresponding perturbations.
Kaggle's Religious and Philosophical texts dataset \cite{kagglereligion} is used as a corpus to identify the religion words. This dataset consists of five texts taken from Project Gutenberg, which is an online archive of religious books. The five textbooks are 1. The King James Bible, 2. The Quran, 3. The Book Of Mormon, 4. The Gospel of Buddha, and 5. Meditations, by Marcus Aurelius. The data consists of 59722 text documents. All documents as a chunk are lemmatized using nltk's WordNet Lemmatizer followed by stemming using Snowball Stemmer. To generate the vocabulary, 100 most frequent words are identified followed by removing words with length less than 3. We end up with 54 religion-related words. Perturbations were manually curated by analyzing and contextualizing each word from Wikipedia's page on Religion \cite{wiki:religion} and then, identifying corresponding perturbations. Table \ref{table:mainreligionperturbationwords} provides a subset-resource for religion-related words and their corresponding perturbations curated from Project Gutenberg's religious books. For generating counterfactuals, religious words can be perturbed to any of its corresponding perturbations chosen randomly. The complete resource is detailed in the Appendix.

\begin{table}[h!]
\centering
\small
\begin{tabular}{|c|c|}
    \hline
    \textbf{Words} & \textbf{Perturbations}\\
    \hline
    christians & hindus,muslims,sikhs,jews\\
    muslims & hindus,sikhs,jews,christians\\
    hindus & muslims,sikhs,jews,christians\\
    jews & hindus,muslims,sikhs,christians\\
    church & mosque,temple,synagogue\\
    mosque & church,temple,gurudwara,synagogue,monastery\\
    temple & church,mosque,gurudwara,synagogue,monastery\\
    bible & quran,gita,tipitaka,granth-sahib,torah\\
    gita & bible,quran,tipitaka,granth-sahib,torah\\
    god & devil\\
    vishnu & jesus,muhammad,moses,buddha\\
    buddha & jesus,muhammad,moses,vishnu\\
    rome & mecca,israel,varanasi\\
    holy & unholy\\
    \hline
\end{tabular}
\caption{\label{table:mainreligionperturbationwords}Words and their corresponding perturbations for sensitive attribute \textit{religion}}
\end{table}

\section{Datasets} \label{data}
We have established our propositions and validated them on three widely used datasets in the area of Fairness in Text.
\subsubsection{Jigsaw Unintended Bias in Toxicity Classification} 
It is the Wikipedia Talk Page Dataset \cite{borkan2019nuanced} that is publicly available at Kaggle \cite{kagglejigsaw} and prepared by Jigsaw \cite{dixon2018measuring}. The data consists of instances of 226235 online comments. These comments are manually labeled into various levels of \textit{TOXICITY}. Apart from descriptive multi-label toxicity labels, there is another \textit{target} column with binary class which signify where a comment text is \textit{TOXIC} or \textit{NON-TOXIC}. 
\subsubsection{IMDB Dataset of 50K Movie Reviews}
It is the Large Movie Review Dataset \cite{maas-EtAl:2011:ACL-HLT2011} with 50,000 movie reviews labeled with the binary sentiment of positive and negative. The dataset contains an even number of positive and negative reviews. It is publicly available at Kaggle \cite{kaggleimdb}.
\subsubsection{SMS Spam Collection Dataset}
It is a set of SMS tagged messages that contain a set of 5,574 SMS messages in English tagged accordingly being ham (legitimate) or spam \cite{almeida2011contributions,hidalgo2012validity,almeida2013towards,webspam}. It is publicly available at Kaggle \cite{kagglespam}.

\section{Experiment}
The details of the models used in the experiments are: 1) Logistic Regression: sklearn linear model is used with lbfgs solver, L2 penalty, while rest are default parameters. 2) Neural Network: The sequential model consists of three hidden layers with units = [12, 8, 6]. The input dimension is 32 and relu activation function is used. 3) Naïve-Bayes: sklearn Gaussian Naïve Bayes model is used with all parameters set to default. 

\subsection{Counterfactual Bias detection}
We perform \textit{Counterfactual Bias} detection on the three datasets using commonly used classification models and compare results between counterfactual single-token and multi-token generation methods. The experiments are performed for each sensitive attribute individually and as one case altogether. Classification models used include an embedding layer.
\subsubsection{Jigsaw Dataset}
Model: Logistic regression. Results: Table \ref{table:jigsawflip}. Accuracy: 92.31\%
\begin{table}[h!]
\small
\makebox[\linewidth]{
\begin{tabular}{|c|c|c|}
    \hline
    \textbf{attribute} & \textbf{single-token} &\textbf{multi-token}\\
    \hline
    age&3.25&4.56\\
    disability&2.56&3.98\\
    race&4.47&6.83\\
    nat&2.37&5.62\\
    gender&0.94&1.38\\
    religion&1.93&2.33\\
    all&3.11&4.38\\
    \hline
\end{tabular}
}
\caption{\label{table:jigsawflip}Flip-rate comparison on jigsaw dataset}
\end{table}  

\subsubsection{IMDB Dataset}
Model: Neural Network. Results: Table \ref{table:imdbflip}. Accuracy: 90.78\%
\begin{table}[h!]
\small
\makebox[\linewidth]{
\begin{tabular}{|c|c|c|}
    \hline
    \textbf{attribute} & \textbf{single-token} &\textbf{multi-token}\\
    \hline
    age&3.13&4.62\\
    disability&2.46&4.31\\
    race&4.38&7.2\\
    nat&4.22&6.7\\
    gender&0.82&1.42\\
    religion&1.74&2.4\\
    all&2.9&4.73\\
    \hline
\end{tabular}
}
\caption{\label{table:imdbflip}Flip-rate comparison on imdb dataset}
\end{table}  

\subsubsection{SPAM Dataset}
Model: Naive-Bayes. Results: Table \ref{table:spamflip}. Accuracy: 89.46\%
\begin{table}[h!]
\small
\makebox[\linewidth]{
\begin{tabular}{|c|c|c|}
    \hline
    \textbf{attribute} & \textbf{single-token} &\textbf{multi-token}\\
    \hline
    age&0.2&0.25\\
    disability&0.4&1.1\\
    race&0&0.16\\
    nat&0&0.2\\
    gender&0.11&0.23\\
    religion&0&0.15\\
    all&0.36&0.65\\
    \hline
\end{tabular}
}
\caption{\label{table:spamflip}Flip-rate comparison on spam dataset}
\end{table}  

Based on the results, the multi-token generation outperforms the single-token generation method.
\subsection{Performance improvement using Explainability}
The experimental results using the two explainability methods on the datasets are in Table \ref{table:jigsawex}, \ref{table:imdbex}, and \ref{table:spamex}.
\begin{table}[h!]
\small
\makebox[\linewidth]{
\begin{tabular}{|c|c|c|c|}
    \hline
    \textbf{attribute} & \textbf{multi-token} &\textbf{multi-token}& \textbf{multi-token}\\
    & & \textbf{and LIME}& \textbf{and Anchors}\\
    \hline
    age&4.56&4.58&4.63\\
    disability&3.98&4.03&4.08\\
    race&6.83&7.05&7.25\\
    nat&5.62&7.15&7.24\\
    gender&1.38&1.45&1.50\\
    religion&2.33&2.35&2.41\\
    all&4.38&4.41&4.52\\
    \hline
\end{tabular}
}
\caption{\label{table:jigsawex}Flip-rate comparison on jigsaw dataset using explainability}
\end{table}  

\begin{table}[h!]
\small
\makebox[\linewidth]{
\begin{tabular}{|c|c|c|c|}
    \hline
    \textbf{attribute} & \textbf{multi-token} &\textbf{multi-token}& \textbf{multi-token}\\
    & & \textbf{and LIME}& \textbf{and Anchors}\\
    \hline
    age&4.62&4.64&4.71\\
    disability&4.31&4.33&4.48\\
    race&7.2&7.32&7.74\\
    nat&6.7&7.2&7.4\\
    gender&1.42&1.44&1.52\\
    religion&2.4&2.41&2.45\\
    all&4.73&4.75&4.82\\
    \hline
\end{tabular}
}
\caption{\label{table:imdbex}Flip-rate comparison on imdb dataset using explainability}
\end{table}  

\begin{table}[h!]
\small
\makebox[\linewidth]{
\begin{tabular}{|c|c|c|c|}
    \hline
    \textbf{attribute} & \textbf{multi-token} &\textbf{multi-token}& \textbf{multi-token}\\
    & & \textbf{and LIME}& \textbf{and Anchors}\\
    \hline
    age&0.25&0.25&0.4\\
    disability&1.1&1.15&1.23\\
    race&0.16&0.16&0.2\\
    nat&0.2&0.21&0.24\\
    gender&0.23&0.23&0.25\\
    religion&0.15&0.16&0.23\\
    all&0.65&0.67&0.7\\
    \hline
\end{tabular}
}
\caption{\label{table:spamex}Flip-rate comparison on spam dataset using explainability}
\end{table}  

From the results, we can infer that ability to generate flipped counterfactuals are higher when we use explainability token(s) along with sensitive token(s). Also, out of the two explainability methods, we can observe that Anchors is performing better than LIME. As mentioned in \cite{ribeiro2018anchors}, Anchors generate high precision and complete coverage explanation. Anchors lead to higher human precision than linear explanations. Using Anchors token(s) along with sensitive token(s), we can generate a higher number of flipped counterfactuals. Hence, for experimentation purposes in the following sub-section, we will be using anchors token(s) along with sensitive token(s) in the multi-token setup.

\subsection{Counterfactual Bias Mitigation}
Following is the experimental method for data augmentation based bias-mitigation to achieve Counterfactual Multi-token Fairness:
\begin{enumerate}
    \item Each sensitive attribute-specific data is split into an 80:20 ratio for training and testing, respectively. The same process is followed for the entire data for performing the experiment combining all sensitive attributes. 
    \item Sensitive attribute-specific data includes only those texts in which that particular sensitive attribute token(s) are present. 
    \item Now, the following steps will be performed for both single-token and multi-token setup (on entire data and each attribute specific data) and, results
    will be compared. 
    \item We also have one setup where after re-training based on the multi-token method, the flip-rate is calculated on the test data using the single-token method. This will signify how the augmentation method using the multi-token setup can improve the overall fairness proposition.
    \begin{enumerate}
        \item The model is trained on the training data.
        \item The trained data is perturbed to generate counterfactuals using the sensitive token(s) and their corresponding perturbation token(s).
        \item Flipped counterfactuals are identified.
        \item Using the trained model, flip-rate is calculated on the test data.
        \item The flipped samples are tagged the same label as original texts and is augmented with the training data.
        \item The combined data is used to re-train the model.
        \item Using the re-trained model, the flip-rate in re-calculated on the test data.
        \item Here, we define \textbf{\textit{Counterfactual Fairness Increment (CFI)}} (in \%) equal to \textit{flip-rate decrement} (in \%).
        \item Calculation of flip-rate using the single-token method on the multi-token method re-trained model is also performed.
    \end{enumerate}
    \item We compare \textit{Counterfactual Fairness Increment (CFI)} (in \%) between the single-token and multi-token method for the entire data and each attribute. Along-with, we analyse the \textbf{Fairness-Utility trade-off} and identify the \textit{Accuracy Drop} (AD in \%) post re-training.
\end{enumerate}

Based on the above experimental method, results are obtained for all three datasets.

\begin{table*}[h!]
\small
\makebox[\linewidth]{
\begin{tabular}{|c|c|c|c|c|c|c|c|c|c|}
    \hline
    \textbf{attribute} & \multicolumn{4}{|c|}{\textbf{single-token fr\%}}
    &\multicolumn{4}{|c|}{\textbf{multi-token fr\%}}&\textbf{single-token fr\% on}\\
    \cline{2-9}
    &&&&&&&&&\textbf{multi-token}\\
    &pre-retraining&post-retraining&AD&\textbf{CFI}&pre-retraining&post-retraining&AD&\textbf{CFI}&\textbf{re-training method}\\
    \hline
    age&3.34&2.98&2.4&10.78&4.63&3.0&2.4&35.2&2.64\\
    disability&2.91&2.12&3.1&27.1&4.07&2.41&2.8&40.8&1.74\\
    race&4.53&3.83&2.3&15.4&7.31&4.27&2.3&41.6&3.2\\
    nat&4.3&3.4&2.4&20.9&7.11&4.17&2.2&41.4&3.0\\
    gender&0.99&0.67&2.4&32.3&1.52&0.7&2.5&54&0.53\\
    religion&2.07&1.3&2.5&37.2&2.42&1.5&2.6&38&0.9\\
    all&3.13&2.75&3.4&12.14&4.55&2.77&3.2&39.12&2.4\\
    \hline
\end{tabular}
}
\caption{\label{table:jigsawfair}Counterfactual Fairness Increment comparison on jigsaw dataset}
\end{table*}  

\begin{table*}[h!]
\small
\makebox[\linewidth]{
\begin{tabular}{|c|c|c|c|c|c|c|c|c|c|}
    \hline
    \textbf{attribute} & \multicolumn{4}{|c|}{\textbf{single-token fr\%}}
    &\multicolumn{4}{|c|}{\textbf{multi-token fr\%}}&\textbf{single-token fr\% on}\\
    \cline{2-9}
    &&&&&&&&&\textbf{multi-token}\\
    &pre-retraining&post-retraining&AD&\textbf{CFI}&pre-retraining&post-retraining&AD&\textbf{CFI}&\textbf{re-training method}\\
    \hline
    age&3.15&3.1&1.2&1.6&4.8&3.52&1.8&26.67&2.85\\
    disability&2.5&2.3&2.1&8&4.5&2.47&2&45.11&2.15\\
    race&4.3&3.5&2.1&18.6&7.8&3.8&2.1&51.28&3.4\\
    nat&4.1&3.4&2&17.1&7.2&3.6&2&50&3.2\\
    gender&1&0.8&2&20&1.5&0.85&2.1&43.33&0.6\\
    religion&1.95&1.62&2.4&16.92&2.5&1.75&2.3&30&1.5\\
    all&3&2.5&2.6&16.67&4.9&3.2&2.7&34.7&2.3\\
    \hline
\end{tabular}
}
\caption{\label{table:imdbfair}Counterfactual Fairness Increment comparison on imdb dataset}
\end{table*}  
\begin{table*}[h!]
\small
\makebox[\linewidth]{
\begin{tabular}{|c|c|c|c|c|c|c|c|c|c|}
    \hline
    \textbf{attribute} & \multicolumn{4}{|c|}{\textbf{single-token fr\%}}
    &\multicolumn{4}{|c|}{\textbf{multi-token fr\%}}&\textbf{single-token fr\% on}\\
    \cline{2-9}
    &&&&&&&&&\textbf{multi-token}\\
    &pre-retraining&post-retraining&AD&\textbf{CFI}&pre-retraining&post-retraining&AD&\textbf{CFI}&\textbf{re-training method}\\
    \hline
    age&0.2&0.18&1.1&10&0.42&0.24&1.2&42.86&0.15\\
    disability&0.5&0.4&1.5&20&1.25&0.71&1.4&43.2&0.36\\
    race&0.1&0.1&1.4&0&0.2&0.1&1.6&50&0\\
    nat&0.1&0.1&2.1&0&0.3&0.1&2.1&66.67&0\\
    gender&0.13&0.1&1.1&23.08&0.27&0.12&1.6&55.56&0.07\\
    religion&0.2&0.08&2.1&60&0.25&0.1&1.9&60&0\\
    all&0.34&0.3&2.1&11.76&0.8&0.4&2.3&50&0.25\\
    \hline
\end{tabular}
}
\caption{\label{table:spamfair}Counterfactual Fairness Increment comparison on spam dataset}
\end{table*}

We can draw the following inferences from tables \ref{table:jigsawfair}, \ref{table:imdbfair}, and \ref{table:spamfair}:
\begin{enumerate}
    \item \textit{Counterfactual Fairness Increment} (in \%) is significantly higher in the multi-token method than the single-token method. Improvement in the \textit{CFI} across all datasets and sensitive attributes show that the multi-token method works better than the single-token method.
    \item We observe the least flip-rate in the overall experiment setup when the flip-rate calculation is done using the single-token method on the multi-token method re-trained model. This signifies that the \textbf{Counterfactual Multi-token Fairness} performs best in both direct and indirect setup.
    \item A max \textit{accuracy} drop of around 3\% is noted post retraining while an average \textit{fairness} rise of 30-40\%, and even 60\% in some cases is observed. So, one has to handle a trade-off of 3\% fall in \textit{accuracy} to gain a high percentage \textit{fairness} increment. Also, accuracy drop difference between single-token and multi-token method is negligible.
\end{enumerate}

\section{Conclusion and Future Work}

This work presents a novel data collection and curation of sensitive tokens and identification of their corresponding perturbation token(s) across multiple sensitive attributes like age, disability, race, nationality, gender, and religion.
We present a resource of sensitive words along with their corresponding perturbations that eventually help in counterfactuals generation. We present a detailed data curation for Religion as a sensitive attribute by coming up with its words and corresponding perturbations.
We have showcased a method for identifying multiple sensitive tokens in texts which extend beyond small and single sentence analysis. These multiple tokens can belong to a particular sensitive attribute or can be distributed across multiple sensitive attributes. Using the resource of sensitive tokens and their perturbations, we have proposed a novel multi-token counterfactuals generation method. This method took into consideration contextual, dependency information, and grammatical understanding while generating counterfactuals. As part of this work, we have used third-party contextual and grammar checking solutions. However, as part of the future work, we aim to cover the full or semi-contextual understanding of texts while generating counterfactuals with the multi-token setup. Some work has already been done in this field in \cite{ebrahimi2017hotflip} and \cite{ribeiro2020beyond}, but they are limited to small and single sentences. With the multi-token setup proposed by us in place, a new dimension of future work opens up. The proposition of methods for generating factually correct counterfactuals will be taken in future work. We have showcased Fairness-Utility trade-off as well. To improve the performance of counterfactuals generation, we have used standard explainability methods. Our proposition gets justified by the improvement in flip-rate scores over single-token and naive multi-token methods. Finally, we perform bias-mitigation to achieve counterfactual multi-token fairness and have shown that fairness achieved using the multi-token method is much higher than the single-token method. 
\clearpage
\bibliographystyle{unsrt}
\bibliography{main}

\clearpage

\appendix

\section{Sensitive Attributes' Words and Perturbations} \label{topic-words}
Counterfactuals are generated by perturbing sensitive words in the original text with their corresponding perturbations. Hence, there is a need for a resource of these words and perturbations for sensitive attributes. We have curated words and corresponding perturbations for the following sensitive attributes. The perturbations are identified by querying the antonyms of sensitive words from Merriam-Webster's Thesaurus \cite{merriam1978merriam}. 

\subsection{AGE}
To identify age-related words, we curate a list of age-related terms, as mentioned in \cite{agelist}. It categorizes words into two divisions i.e., \textit{old} and \textit{young}. Extending the list, we consider seed words in each division and extend them using distance in word embedding space. In Table \ref{table:ageperturbationwords} complete resource of age-related words and their corresponding perturbations have been articulated.
\begin{table}[h!]
\small
\begin{tabular}{|c|c|}\hline
  \textbf{Topic words} & \textbf{Perturbation words}\\  
  \hline
  parents & children\\
  \hline
  elders & kids,teens,teenagers\\
  \hline
  elderly & kiddish\\
  \hline
  adult & youngster,youth\\
  \hline
  old & teenage\\
  \hline
  elder & teenager\\
  \hline
  parent & child\\
  \hline
  old & young,younger,youngest\\
  \hline
  older & younger\\
  \hline
  oldest & youngest\\
  \hline
  young & old,older,oldest\\
  \hline
  children & parents\\
  \hline
  kids & elders\\
  \hline
  kiddish & elderly\\
  \hline
  teens & elders\\
  \hline
  youngster & adult\\
  \hline
  teenage & old\\
  \hline
  teenager & elder\\
  \hline
  teenagers & elders\\
  \hline
  child & parent\\
  \hline
  youth & adult\\
  \hline
  younger & older\\
  \hline
  youngest & oldest\\
  \hline
\end{tabular}
\caption{\label{table:ageperturbationwords}Topic words and their corresponding perturbation words for sensitive attribute \textit{age}}
\end{table}

\subsection{DISABILITY}
Social Biases as barriers for a person with disabilities have been studied in \cite{disabilitywords}. It highlights terms or words used in texts when representing disability. 
Disability-related words are queried, and their antonyms are identified as corresponding perturbations. Table \ref{table:disabilityperturbationwords} provides a complete resource for disability-related words and their corresponding perturbations.
\begin{table}[h!]
\small
\begin{tabular}{|c|c|}\hline
    \textbf{Topic words} & \textbf{Perturbation words}\\  
    \hline
    mental & sane\\
    \hline
    mentally & intellectually\\
    \hline
    illness & healthy\\
    \hline
    blind & sighted\\
    \hline
    deaf & hearing\\
    \hline
    sight-deficient & sighted\\
    \hline
    depression & cheerfulness\\
    \hline
    insane &  sane\\
    \hline
    dyslexia &  unimpaired\\
    \hline
    slow-learner &  fast-leaner\\
    \hline
    losers &  winners\\
    \hline
    loser & winner\\
    \hline
    slow & fast\\
    \hline
    depressed & cheerful\\
    \hline
    disorder & in-order\\
    \hline
    ill & healthy\\
    \hline
    problems & solutions\\
    \hline
    issues & solutions\\ 
    \hline
    help & no-help\\
    \hline
    treatment & no-treatment\\
    \hline
    care & neglect\\
    \hline
    medication & hindrance\\
    \hline
    therapy & inattention\\
    \hline
    treated & mistreated\\
    \hline
    counseling & discourage\\
    \hline
    meds &    diseases\\
    \hline
    medications & hindrance\\
    \hline
    homeless & settled\\
    \hline
    drugs & diseases\\
    \hline
    homelessness & settleness\\
    \hline
    drug & disease\\
    \hline
    alcohol & non-intoxicant\\
    \hline
    police & civilian\\
    \hline
    addicts & non-addicts\\
    \hline
\end{tabular}
\caption{\label{table:disabilityperturbationwords}Topic words and their corresponding perturbation words for sensitive attribute \textit{disability}}
\end{table}

\subsection{RACE}
As a sensitive attribute, Race has been studied thoroughly in standard datasets like UCI \cite{Dua:2019} Adult (an income dataset based on a 1994 US Census database; sensitive attributes: sex, race), and ProPublica COMPAS (a prison recidivism dataset; sensitive attributes: sex, race). We have compiled race-related words from these two datasets in the following list: [\textit{black, white, asian, caucasian, hispanic, asian-pac-islander, amer-indian-eskimo, arabic, and oriental}]. Perturbations for \textit{Race} can be any of the other words present in their respective lists. Table \ref{table:fullraceperturbationwords} shows a subset of race related words and their corresponding perturbations from which one word can be randomly chosen as its perturbation for generating counterfactuals.

\begin{table}[h!]
\centering
\small
\begin{tabular}{|c|c|}\hline
        \textbf{Topic words} & \textbf{Perturbation words}\\
        \hline
        black & white,asian,caucasian,\\
        &hispanic,asian-pac-islander,\\
        &amer-indian-eskimo,arabic,oriental\\
        \hline
        white & black,asian,caucasian,\\
        &hispanic,asian-pac-islander,\\
        &amer-indian-eskimo,arabic,oriental\\
        \hline
        asian & black,white,caucasian,\\
        &hispanic,asian-pac-islander,\\
        &amer-indian-eskimo,arabic,oriental\\
        \hline
        British & Canadians,Indians\\
        &Americans,Chinese,other nationalities\\
        \hline
        Chinese & Canadians,Indians\\
        &Americans,Russians,other nationalities\\
        \hline
        Americans & Canadians,Indians\\
        &Chinese,Russians,other nationalities\\
        \hline
\end{tabular}
\caption{\label{table:fullraceperturbationwords}Words and their corresponding perturbations for sensitive attribute \textit{race}}
\end{table}

\subsection{NATIONALITY}
Nationality: We have curated 222 nationality-related words from the Wikipedia list of nationalities \cite{wiki:NAT}. Perturbations for \textit{Nationality} can be any of the other words present in their respective lists. Table \ref{table:fullnatperturbationwords} shows a subset of nationality related words and their corresponding perturbations from which one word can be randomly chosen as its perturbation for generating counterfactuals.

\begin{table}[h!]
\centering
\small
\begin{tabular}{|c|c|}\hline
        \textbf{Words} & \textbf{Perturbations}\\
        \hline
        British & Canadians,Indians\\
        &Americans,Chinese,other nationalities\\
        \hline
        Chinese & Canadians,Indians\\
        &Americans,Russians,other nationalities\\
        \hline
        Americans & Canadians,Indians\\
        &Chinese,Russians,other nationalities\\
        \hline
\end{tabular}
\caption{\label{table:fullnatperturbationwords}Words and their corresponding perturbations for sensitive attribute \textit{nat}}
\end{table}

\subsection{GENDER}
To identify gender words, we take the list described in \cite{bolukbasi2016man}. We have considered gender-specific keywords extracted using word embeddings described in \cite{bolukbasi2016man} totaling 165-word replacements. These word replacements are perturbations words. 
The complete resource containing gender words and their corresponding perturbations are articulated in Table \ref{table:genderperturbationwords}.

\begin{table*}[h!]
\small
\makebox[\linewidth]{
\begin{tabular}{|c|c|c|c|c|c|}
    \hline
    \textbf{Topic words} & \textbf{Perturbation words} & \textbf{Topic words} & \textbf{Perturbation words} & \textbf{Topic words} & \textbf{Perturbation words}\\
    \hline
    
    he & she &
    he's & she's &
    He's & She's \\
    his & her &
    her & his &
    she & he \\
    she's & he's &
    She's & He's &
    him & her \\
    man & woman &
    women & men &
    men & women \\
    woman & man &
    spokesman & spokeswoman &
    wife & husband \\
    himself & herself &
    son & daughter &
    mother & father \\
    father & mother &
    chairman & chairwoman &
    daughter & son \\
    husband & wife &
    guy & girl &
    girls & guys \\
    girl & guy &
    boy & girl &
    boys & girls \\
    brother & sister &
    spokeswoman & spokesman &
    female & male \\
    sister & brother &
    male & female &
    herself & himself \\
    brothers & sisters &
    dad & mom &
    actress & actor \\
    mom & dad &
    sons & daughters &
    girlfriend & boyfriend \\
    daughters & sons &
    lady & gentleman &
    boyfriend & girlfriend \\
    sisters & brothers &
    mothers & fathers &
    king & queen \\
    businessman & businesswoman &
    grandmother & grandfather &
    grandfather & grandmother \\
    ladies & gentlemen &
    uncle & aunty &
    males & females \\
    congressman & congresswoman &
    grandson & granddaugther &
    queen & king \\
     businessmen & businesswomen &
    wives & husbands &
    widow & widower \\
    nephew & niece &
    bride & bridegroom &
    females & males \\
    aunt & uncle &
    gay & lesbian &
    gays & lesbians \\
    lesbian & gay &
    chairwoman & chairman &
    fathers & mothers \\
    moms & dads &
    maiden & master &
    bisexual & heterosexual \\
    granddaughter & grandson &
    younger-brother & younger-sister &
    lads & lass \\
    lion & lioness &
    bisexuals & heterosexuals &
    gentleman & gentlewoman \\
    homosexual & heterosexual &
    homosexuals & heterosexuals &
    bachelor & bachelorette \\
    niece & nephew &
    husbands & wives &
    prince & princess \\
    salesman & saleswoman &
    hers & hims &
    dude & dudette \\
    princess & prince &
    lesbians & gays &
    councilman & councilwoman \\
    actresses & actors &
    gentlemen & gentlewomen &
    stepfather & stepmother \\
    monks & nuns &
    ex-girlfriend & ex-boyfriend &
    lad & lass \\
    nephews & nieces &
    maid & man-servant &
    daddy & mommy \\
    fiance & fiancee &
    fiancee & fiance &
    kings & queens \\
    dads & moms &
    waitress & waiter &
    maternal & paternal \\
    heroine & hero &
    nieces & nephews &
    girlfriends & boyfriends \\
    sir & madam &
    mistress & master &
    grandma & grandpa \\
    widows & widowers &
    diva & divus &
    teenage-boy & teenage-girl \\
    nuns & monks &
    countrymen & countrywomen &
    teenage-girl & teenage-boy \\
    nun & monk &
    brides & bridegrooms &
    housewife & househusband \\
    spokesmen & spokeswomen &
    suitors & female-suitors &
    motherhood & fatherhood \\
    stepmother & stepfather &
    hostess & host &
    schoolboy & schoolgirl \\
    brotherhood & sisterhood &
    stepson & stepdaughter &
    congresswoman & congressman \\
    uncles & aunties &
    witch & wizard &
    monk & nun \\
    paternity & maternity &
    suitor & female-suitor &
    businesswoman & businessman \\
    gal & guy &
    statesman & stateswoman &
    schoolgirl & schoolboy \\
    fathered & mothered &
    goddess & god &
    stepdaughter & stepson \\
    grandsons & granddaughters &
    godfather & godmother &
    mommy & daddy \\
    boyhood & girlhood &
    grandmothers & grandfathers &
    grandpa & grandma \\
    boyfriends & girlfriends &
    queens & kings &
    witches & wizards \\
    aunts & uncles &
    granddaughters & grandsons &
    heterosexual & homosexual \\
    heterosexuals & homosexuals &
    widower & widow &
    salesmen & saleswomen \\
    maids & man-servants &
    gals & guys &
    housewives & househusbands \\
    fatherhood & motherhood &
    princes & princesses &
    matriarch & patriarch \\
    patriarch & matriarch &
    ma & pa &
    pa & ma \\
    councilmen & councilwomen &
    mr & mrs &
    mrs & mr \\
    \hline
    
\end{tabular}
}
\caption{\label{table:genderperturbationwords}Topic words and their corresponding perturbation words for sensitive attribute \textit{gender}}
\end{table*}

\subsection{RELIGION}
Till now religion as a sensitive attribute hasn't been analyzed in depth. One of the novel aspects of this paper is the identification of religion-related words and their corresponding perturbations.
Kaggle's Religious and Philosophical texts dataset \cite{kagglereligion} is used as a corpus to identify the religion words. This dataset consists of five texts taken from Project Gutenberg, which is an online archive of religious books. The five textbooks are 1. The King James Bible, 2. The Quran, 3. The Book Of Mormon, 4. The Gospel of Buddha, and 5. Meditations, by Marcus Aurelius. The data consists of 59722 text documents. All documents as a chunk are lemmatized using nltk's WordNet Lemmatizer followed by stemming using Snowball Stemmer. To generate the vocabulary, 100 most frequent words are identified followed by removing words with length less than 3. We end up with 54 religion-related words. Perturbations were manually curated by analyzing and contextualizing each word from Wikipedia's page on Religion \cite{wiki:religion} and then, identifying corresponding perturbations. Table \ref{table:religionperturbationwords} provides a complete resource for religion-related words and their corresponding perturbations curated from Project Gutenberg's religious books. For generating counterfactuals, religious words can be perturbed to any of its corresponding perturbations chosen randomly.

\begin{table}[h!]
\small
\begin{tabular}{|c|c|}
    \hline
    \textbf{Topic words} & \textbf{Perturbation words}\\
    \hline
    muslims & hindus,sikhs,jews,christians,buddhists,atheists\\
    hindus & muslims,sikhs,jews,christians,buddhists,atheists\\
    sikhs & hindus,muslims,jews,christians,buddhists,atheists\\
    jews & hindus,muslims,sikhs,christians,buddhists,atheists\\
    christians & hindus,muslims,sikhs,jews,buddhists,atheists\\
    buddhists & hindus,muslims,sikhs,jews,christians,atheists\\
    atheists & hindus,muslims,sikhs,jews,christians,buddhists\\
    church & mosque,temple,gurudwara,synagogue,monastery\\
    cathedral & mosque,temple,gurudwara,synagogue,monastery\\
    basilica & mosque,temple,gurudwara,synagogue,monastery\\
    chapel & mosque,temple,gurudwara,synagogue,monastery\\
    mosque & church,temple,gurudwara,synagogue,monastery\\
    masjid & church,temple,gurudwara,synagogue,monastery\\
    temple & church,mosque,gurudwara,synagogue,monastery\\
    gurudwara & church,temple,mosque,synagogue,monastery\\
    synagogue & church,temple,gurudwara,mosque,monastery\\
    monastery & church,temple,gurudwara,synagogue,mosque\\
    bible & quran,gita,tipitaka,granth-sahib,torah\\
    quran & bible,gita,tipitaka,granth-sahib,torah\\
    gita & bible,quran,tipitaka,granth-sahib,torah\\
    ramayana & bible,quran,tipitaka,granth-sahib,torah\\
    mahabharata & bible,quran,tipitaka,granth-sahib,torah\\
    tipitaka & bible,quran,gita,granth-sahib,torah\\
    granth-sahib & bible,quran,gita,tipitaka,torah\\
    torah & bible,quran,gita,tipitaka,granth-sahib\\
    vedas & bible,quran,tipitaka,granth-sahib,torah\\
    testament & quran,gita,tipitaka,granth-sahib\\
    talmud & bible,quran,gita,tipitaka,granth-sahib\\
    god & devil\\
    devil & god,lord,allah,prophet\\
    lord & devil\\
    allah & devil\\
    jesus & muhammad,moses,vishnu,buddha\\
    prophet & devil\\
    moses & jesus,muhammad,vishnu,buddha\\
    muhammad & jesus,moses,vishnu,buddha\\
    peter & muhammad,moses,vishnu,buddha\\
    john & muhammad,moses,vishnu,buddha\\
    jacob & jesus,muhammad,vishnu,buddha\\
    samuel & jesus,muhammad,vishnu,buddha\\
    christ & muhammad,moses,vishnu,buddha\\
    vishnu & jesus,muhammad,moses,buddha\\
    buddha & jesus,muhammad,moses,vishnu\\
    gospel & devil\\
    israel & mecca,rome,varanasi,bodh-gaya\\
    mecca & israel,rome,varanasi,bodh-gaya\\
    rome & mecca,israel,varanasi,bodh-gaya\\
    medina & israel,rome,varanasi,bodh-gaya\\
    jerusalem & varanasi,bodh-gaya\\
    arabia & israel,rome,varanasi,bodh-gaya\\
    varanasi & israel,rome,mecca,bodh-gaya\\
    bodh-gaya & israel,rome,varanasi,mecca\\
    holy & unholy\\
    unholy & holy\\
    \hline
\end{tabular}
\caption{\label{table:religionperturbationwords}Topic words and their corresponding perturbation words for sensitive attribute \textit{religion}}
\end{table}

\end{document}